\documentclass[10pt,twocolumn,letterpaper]{article}

\usepackage{iccv}
\usepackage{times}
\usepackage{epsfig}
\usepackage{graphicx}
\usepackage{amsmath}
\usepackage{amssymb}

\usepackage[breaklinks=true,bookmarks=false]{hyperref}

\newcommand{\x}{\mathbf{x}}
\newcommand{\X}{\mathbf{X}}
\newcommand{\y}{\mathbf{y}}

\newcommand{\Z}{\mathbf{Z}}
\newcommand{\A}{\mathbf{A}}

\newcommand{\W}{\mathbf{W}}

\iccvfinalcopy % *** Uncomment this line for the final submission

\def\iccvPaperID{****} % *** Enter the ICCV Paper ID here

\ificcvfinal\fi

\begin{document}

%%%%%%%%% TITLE
\title{Polarized Self-Attention: Towards High-quality Pixel-wise Regression}

\author{Huajun Liu$^{12}$\thanks{Work partially done as a postdoc at Carnegie Mellon University.}, Fuqiang Liu$^{2}$, Xinyi Fan$^{1}$ and Dong Huang$^{2}$\\
{\small $^{1}$Nanjing University of Science and Technology,$^{2}$Carnegie Mellon University}\\
{\tt\small  \{liuhj,xyxy\}@njust.edu.cn; fuqiangl@andrew.cmu.edu; donghuang@cmu.edu} 
}

\maketitle
\ificcvfinal\thispagestyle{empty}\fi

%%%%%%%%% ABSTRACT
\begin{abstract}
 Pixel-wise regression is probably the most common problem in fine-grained computer vision tasks, such as estimating keypoint heatmaps and segmentation masks. These regression problems are very challenging particularly because they require, at low computation overheads, modeling long-range dependencies on high-resolution inputs/outputs to estimate the highly nonlinear pixel-wise semantics. While attention mechanisms in Deep Convolutional Neural Networks(DCNNs) has become popular for boosting long-range dependencies, element-specific attention, such as Nonlocal blocks, is highly complex and noise-sensitive to learn, and most of simplified attention hybrids try to reach the best compromise among multiple types of tasks. In this paper, we present the Polarized Self-Attention(PSA) block that incorporates two critical designs towards high-quality pixel-wise regression: (1) Polarized filtering: keeping high internal resolution in both channel and spatial attention computation while completely collapsing input tensors along their counterpart dimensions. (2) Enhancement: composing non-linearity that directly fits the output distribution of typical fine-grained regression, such as the 2D Gaussian distribution (keypoint heatmaps), or the 2D Binormial distribution (binary segmentation masks). PSA appears to have exhausted the representation capacity within its channel-only and spatial-only branches, such that there is only marginal metric differences between its sequential and parallel layouts. Experimental results show that PSA boosts standard baselines by $2-4$ points, and boosts state-of-the-arts by $1-2$ points on 2D pose estimation and semantic segmentation benchmarks. Codes are released\footnote{\url{https://github.com/DeLightCMU/PSA}}.
%   Incorporating attention mechanism in Deep Convolutional Neural Networks(DCNNs) has become a popular practice to handle long-range dependencies in computer vision tasks, while a painful trade-off between performance gain and computation overheads. Adding standalone attention blocks upon a vallina architecture of convolution blocks seems an short-path to boost performance, but directly aggregates the computation, parameter and memory of both blocks. On the other hand, designing a brand new attention-ware DCNN architectures require a tedious process of searching network-dependent training tricks and hyper-parameters.  
\end{abstract}

%%%%%%%%% BODY TEXT
\section{Introduction}
\begin{figure}[!htb]
\centering
\includegraphics[width=.99\linewidth]{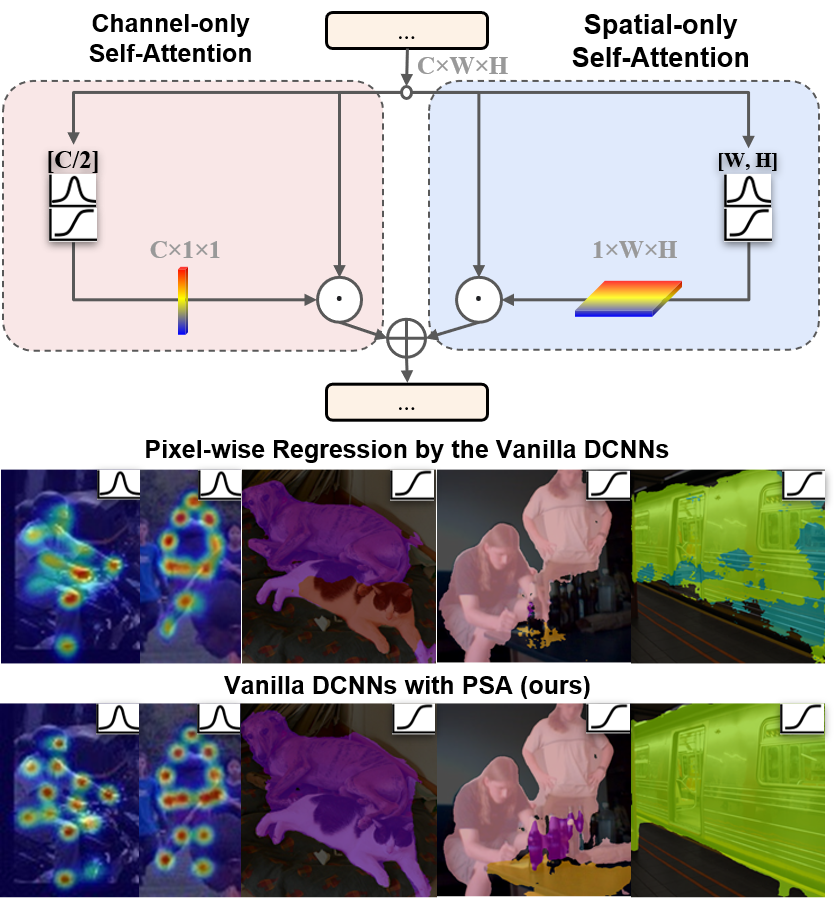}
\caption{\textbf{Polarized Self-Attention(PSA) block (ours)}, keeps high internal resolution along the channel ($C/2$) and spatial dimension($[W,H]$) while collapses the input tensor $[C\times W\times H]$ along their counterparts dimensions, and fits output distributions of pixel-wise regression with a softmax-sigmoid composition. At minor computation-memory overheads upon the vanilla DCNNs, PSA produces significantly higher-quality person keypoint heatmaps and semantic segmentation masks (also see Table~\ref{table:COCO_Pose}-\ref{table:VOC_Semantic} for the boosts in metrics). }
  \label{fig:Intro}
\end{figure}
Recent trends from the coarse-grained (such as image-wise classification~\cite{Russakovsky15ImageNet} and bounding box detection~\cite{Girshick15FastR-CNN}) to the fine-grained computer vision tasks (such as keypoint estimation~\cite{Luo21} and segmentation segmentation~\cite{Zhong20}) have received booming advances in both research and industrial communities. Comparing to the coarse-grained tasks, perception at the pixel-wise level is increasingly appealing in autonomous driving ~\cite{Treml16}, augment reality~\cite{Chiu18}, medical image processing~\cite{Litjens17}, and public surveillance~\cite{Wang21}. 

The goal of the pixel-wise regression problem is to map every image pixels of the same semantics to the same scores. For instance, mapping all the background pixels to 0 and all the foreground pixels to their class indices, respectively. Two typical tasks are keypoint heatmap regression and segmentation mask regression. Most DCNN models for regression problems take an encoder-decoder architecture. The encoder usually consists of a backbone network, such as ResNet~\cite{He16}, that sequentially reduces the spatial resolution and increases the channel resolution, while the decoder usually contains de-convolution/up-sampling operations that recover the spatial resolution and decrease the channel resolution. Typically the tensor connecting the encoder and decoder has an element number smaller than both the input image tensor and the output tensor. The reduction of elements is necessary for computation/memory efficiency and stochastic optimization reasons~\cite{Goodfellow-et-al-2016}. However, the pixel appearances and patch shapes of the same semantics are highly nonlinear in nature and therefore difficult to be encoded with a reduced number of features. Moreover, high input-output resolutions are preferred for fine details of objects and object parts~\cite{Lin2017,Sun2019,Wang2020}. Comparing to the image classification task where an input image is collapsed to an output vector of class indices, the pixel-wise regression problem has a higher problem complexity by the order of output element numbers. From the model design perspective, the pixel-wise regression problem faces special challenges: (1) Keeping high internal resolution at a reasonable cost; (2) Fitting output distribution such as that of the keypoint heatmaps or segmentation masks.  

Based on the tremendous success in new DCNNs architectures, we focus on a plug-and-play solution that could consistently improve an existing (vanilla) network, i.e., inserting attention blocks~\cite{Vaswani17}\cite{Shaw18}\cite{Cordonnier20}\cite{Wang18nonlocal}\cite{Dai19}\cite{Xia19}\cite{Hu18b}\cite{Cao19}. Most of above hybrids try to reach the best compromise among multiple types of tasks, for instance, image classification, object detection, as well as for instance segmentation. These generalized goals are partially the reason that channel-only attention (SE~\cite{Hu18}, GE~\cite{Hu18b} and GCNet~\cite{Cao19}) are among the most popular blocks. Channel-only attention blocks put the same weights on different spatial locations, such that the classification task still benefits since its spatial information eventually collapses by pooling, and the anchor displacement regression in object detection benefits since the channel-only attention unanimously highlights all foreground pixels. Unfortunately, due to critical differences in attention designs, the channel-spatial compositional attention blocks, (e.g., DA~\cite{Fu2019}, CBAM~\cite{Sanghyun2018cbam}), did not show significant overall advantages from the latest channel-only attentions such as GCNet~\cite{Cao19}.  

In this paper, we present the Polarized Self-Attention (PSA) block (See Figure~\ref{fig:Intro}) for high-quality pixel-wise regression. To preserve the potential loss of high-resolution information in vanilla/baseline DCNNs by pooling/downsampling, PSA keeps the highest internal resolution in attention computation among existing attention blocks (see also Table~\ref{table:Analysis}). To fitting the output distribution of typical fine-grained regression, PSA fuse softmax-sigmoid composition in both channel-only and spatial-only attention branches. Comparing to existing channel-spatial compositions~\cite{Sanghyun2018cbam,Fu2019} that favor particular layouts, there is only marginal metric differences between PSA layouts. This indicates PSA may have exhausted the representation capacity within its channel-only and spatial-only branches. We conducted extensive experiments to demonstrate the direct performance gain of PSA on standard baselines as well as state-of-the-arts.

\section{Related Work}

\textbf{Pixel-wise Regression Tasks}: The advances of DCNNs for pixel-wise regression are basically pursuing higher resolution. For body keypoint estimation, Simple-Baseline\cite{Xiao18} consists of conventional components ResNet+deconvolution. HRnet\cite{Sun2019} address the resolution challenge of Simple-Baseline with 4 parallel high-to-low resolution branches and their pyramid fusion. Other most recent variants, DARK-Pose\cite{Zhang2020} and UDP-Pose\cite{Huang2020}, both compensate for the loss of resolution due to the preprocessing, post-processing, and propose techniques to achieve a sub-pixel estimation of keypoints.  Note that, besides the performance gain among network designs, the same models with and $388\times 284$ inputs are usually better than that with $256\times 192$ inputs. This constantly reminds researchers of the importance of keeping high-resolution information. For Semantic segmentation, \cite{Chen17} introduces atrous convolution in the decoder head of Deeplab for wide receptive field on high-resolution inputs. To overcome the limitation of ResNet backbones in Deeplab, all the latest advances are based on HRnet~\cite{Wang2020}, in particular, HRNet-OCR\cite{Tao2020} and its variants are the current state-of-the-art. There are many other multitask architecture ~\cite{He17,zhou2019objects,cheng2020panoptic} that include pixel-wise regression as a component. 

PSA further pursues the high-resolution goals of the above efforts from the attention perspective and further boosts the above DCNNs. 

\textbf{Self-attention and its Variants.}  Attention mechanisms have been introduced into many visual tasks to address the weakness of standard convolutions~\cite{Shaw18}\cite{Bello19}\cite{Andreoli19}\cite{Ramachandran19}\cite{Cao19}. In the self-attention mechanism, each input tensor is used to compute an attention tensor and is then re-weighted by this attention tensor. Self-attention~\cite{Vaswani17}\cite{Shaw18}\cite{Cordonnier20} emerged as a standard component to capture long-range interactions, after it success in sequence modeling and generative modeling tasks. Cordonnier et al.~\cite{Cordonnier20} has proven that a multi-head self-attention layer with a sufficient number of heads is at least as expressive as any convolutional layer. In some vision tasks, such as object detection and image classification, self-attention augmented convolution models~\cite{Bello19} or standalone self-attention models~\cite{Ramachandran19} have yielded remarkable gains. While most self-attention blocks were inserted after convolution blocks, attention-augmented convolution~\cite{Bello19} demonstrates that parallelizing the convolution layer and attention block is a more powerful structure to handle both short and long-range dependency. 

PSA advances self-attention for pixel-wise regression and could also be used in other variants such as the convolution-augmented attentions.

\textbf{Full-tensor and simplified attention blocks.} The basic non-local block (NL)~\cite{Wang18nonlocal}  and its variants, such as a residual form~\cite{Zhang19} second-order non local~\cite{Dai19}\cite{Xia19}, and asymmetric non-local~\cite{Zhu19}, produce full-tensor attentions and have successfully improved person re-identification, image super-resolution, and semantic segmentation tasks. To capture pair-wise similarities among all feature elements, the NL block computes an extremely large similarity matrix between the key feature maps and query feature maps, leading to huge memory and computational costs. EA~\cite{Shen20} produces a low-rank approximation of NL block for computation efficiency.  BAM~\cite{Park18},DAN~\cite{Fu2019} and CBAM~\cite{Sanghyun2018cbam} produce different compositions of the channel-only and spatial-only attentions. Squeeze-and-Excitation (SENet)~\cite{Hu18}, Gather-Excite~\cite{Hu18b} and GCNet~\cite{Cao19} only re-weight feature channels using signals aggregated from global context modeling. Most of above attention blocks were designed as a compromise among multiple types of tasks, and do not address the specific challenges in fine-grained regression. 

PSA address the specific challenges in fine-grained regression by keeping the highest attention resolution among existing attention blocks, and directly fitting the typical output distributions.

 \section{Our Method}
\textbf{Notations:}\footnote{All non-bold letters represent scalars. Bold capital letter $\X$ denotes a matrix; Bold lower-case letters $\bf{x}$ is a column vector. $\x_i$ represents the $i^{th}$ column vector of the matrix $\X$. $x_{j}$ denotes the $j^{th}$ element of $\x$.  $\langle \x, \y \cdot\rangle= \x^{T}\y$ denotes the inner-product between two vectors or metrics.}  Denote $\X \in \Re^{C_{in}\times H\times W \times}$
  as a feature tensor of one sample (e.g., one image), where $C_{in},H,W$ are the number of elements along the height, width, and channel dimension of $\X$, respectively.  
 $\X= \{\x_{i}\}_{i=1}^{HW}$ where $\x_{i}\in \Re^{C_{in}}$ is a feature vector along the channel dimension.  
 A  self-attention block $\A(\cdot)$ takes $\X$ as input, and produces a tensor $\Z$ as output, where $\Z\in \Re^{C_{out}\times H\times W }$. A DCNN block is formulated as a nonlinear mapping $\Psi:\X \rightarrow \Z$. The possible operators of the network block include: the convolution layer $\W(\cdot)$, the batch norm  layer $BN(\cdot)$, the ReLU activation layer $RU(\cdot)$, softmax $SM(\cdot)$. Without losing generality, all the convolution layers in attention blocks are the ($1\times 1$) convolution, denoted by $\W$.  For simplicity, we only consider the case where the input tensor $\X$ and output tensor $\Z$ of a DCNN block have the same dimension $C\times H \times W$  (i.e., $C_{in}=C_{out}$).

\subsection{Self-Attention for Pixel-wise Regression}

A DCNN for pixel-wise regression learns a weighted combination of features along two dimensions: (1) channel-specific weighting to estimate the class-specific output scores; (2) spatial-specific weighting to detect pixels of the same semantics. The self-attention mechanism applied to the DCNN is expected to further highlight features for both above goals. 

Ideally, with a full-tensor self-attention $\Z= \A(\X) \odot \X$ ($\A(\X)\in \Re^{C\times H \times W}$), the highlighting could potentially be achieved at the element-wise granularity ($C\times H \times W$ elements). However, the attention tensor $\A$ is very complex and noise-prone to learn directly. In the Non-Local self-attention block~\cite{Wang18nonlocal}, $\A$ is calculated as,  
\begin{equation}
\A=  \W_{z}(F_{sm}(\X^{T}\W_{k}^{T}\W_{q}\X)\W_{v}\X). 
\end{equation}
There are four ($1\times 1$) convolution kernels, i.e.,  $\W_{z}$,$\W_{k}$, $\W_{q}$, and $\W_{v}$, that learns the linear combination of spatial features among different channels. Within the same channels, the $HW\times HW$ outer-product between $\W_{k}\X$ and $\W_{q}\X$ activates any features at different spatial locations that have a similar intensity. The joint activation mechanism of spatial features is very likely to highlight the spatial noise. The only actual weights, $\W$s, are channel-specific instead of spatial-specific, making the Non-Local attention exceptionally redundant at the huge memory-consumption of the $HW\times HW$ matrix. 
For efficient computation, reduction of NL leads to many possibilities: Low rank approximation of $\A$ (EA), Channel-only self-attention $\A^{ch}\in \Re^{C\times 1 \times 1}$ that highlight the same global context for all pixels(GC~\cite{Cao19} and SE~\cite{Hu18b} ), Spatial-only self-attention $\A^{sp}\in \Re^{1\times W \times H}$ not powerful enough to be recognized as a standalone model, Channel-spatial composition $\A^{sp}$, where the parallel composition: $\Z= \A^{ch} \odot^{ch}\X+ \A^{sp} \odot^{sp} \X $ and the sequential composition: $\Z= \A^{ch} \odot^{ch}(\A^{sp} \odot^{sp}\X)$ introduce different order of non-linearity. Different conclusions were empirically drawn, such as CBAM~\cite{Sanghyun2018cbam}  (sequential$>$parallel) and DA~\cite{Fu2019} (parallel$>$sequential), which partially indicates that the intended non-linearity of the tasks are not fully modeled within the attention blocks. 

These issues are typical examples of general attention design that does not target the pixel-wise regression problem. With the help of Table~\ref{table:Analysis}, we re-visit critical design aspects of existing attention blocks and raise challenges on how to achieve both channel-specific and spatial-specific weighting for pixel-wise regression. (All the attention blocks are compared with their top-performance configurations.) 

\begin{table}[!htb]
\centering
\fontsize{7}{8}\selectfont
\setlength{\tabcolsep}{1.6pt}
\begin{tabular}{l|c|c|c|c}
\hline
\textbf{Method} & ch. resolution& sp. resolution& non-linearity & complexity\hspace{0.1cm} $O(\cdot)$ \\
\hline \hline
NL\cite{Wang18nonlocal} & $C$ &$[W,H]$ & SM & $C^2WH+CW^2H^2$\\  \hline 
GC~\cite{Cao19}   & $C/4$ & - & SM+ReLU & $CWH$ \\ 
SE~\cite{Hu18b}   & $C/4$ & - & ReLU+SD & $CWH$ \\ 
CBAM~\cite{Sanghyun2018cbam} & $C/16$ & $[W,H]$ & SD & $CWH$\\ 
DA~\cite{Fu2019} & $C/8$ & $[W,H]$ & SM & $C^2WH+C W^2 H^2$\\ 
EA~\cite{Shen20} & $d_{k}$ ($\ll C$) & $d_v$ ($\ll min(W,H)$) & SM & $CWH$\\ 
\hline
PSA(ours) & $C/2$& $[W,H]$ & SM+SD & $CWH$\\
\hline
\end{tabular}
\caption{Re-visit critical design aspects in existing attention blocks. All the attention blocks are compared in their top-performance configurations. SM: SoftMax, SD: Sigmoid. Complexity is estimated assuming $C < WH$.}
\label{table:Analysis}
\end{table}

\textbf{Internal Attention Resolution.} Recall that most pixel-wise regression DCNNs use the same backbone networks, e.g., ResNet, as the classification (i.e., image recognition) and coordinate regression(i.e. bbox detection, instance segmentation) tasks. For robustness and computational efficiency, these backbones produce low-resolution features, for instance $1\times 1\times 512$ for the classification and $[W/r,H/r]$ for bbox detection, where $r$ is the longest side pixels of the smallest object bounding box. Pixel-wise regression cannot afford such loss of resolution, especially because the highly non-linearity in object edges and body parts are very difficult to encode in low-resolution features~\cite{Chen17,Wang2020,Sun2019}.

Using these backbones in pixel-wise regression, self-attention blocks are expected to preserve high-resolution semantics in attention computation. However, in Table~\ref{table:Analysis}, all the reductions of NL reach their top performance at a lower internal resolution. Since their performance metrics are far from perfect, the natural question to ask is: \textit{are there better non-linearity that could leverages higher resolution information in attention computation?}

\textbf{Output Distribution/Non-linearity.} In DCNNs for pixel-wise regression, outputs are usually encoded as 3D tensors. For instance, the 2D keypoint coordinates are encoded as a stack of 2D Gaussian maps $[\#keypoint\_type\times W \times H]$. The pixel-wise class indices are encoded as a stack of binary maps $[\#semantic\_classes\times W \times H]$ which follows the Binormial distribution. Non-linearity that directly fits the distribution upon linear transformations (such as convolution) could potentially alleviate the learning burden of DCNNs. The natural nonlinear functions to fit the above distributions are SoftMax for 2D Gaussian maps, and Sigmoid for 2D Binormial Distribution. However, none of the existing attention blocks in Table~\ref{table:Analysis} contains such a combination of nonlinear functions. 

% The 2D Gaussian maps in its distribution density formulation, 
% \begin{equation}
% G(x;\mu, \Sigma)= \frac{1}{2\pi|\Sigma|^{1/2}}exp(-1/2 (x-\mu)^{T}\Sigma^{-1} (x-\mu))  
% \end{equation}
\subsection{Polarized Self-Attention (PSA) Block}
 
\begin{figure}[!htb]
\centering
\includegraphics[width=.99\linewidth]{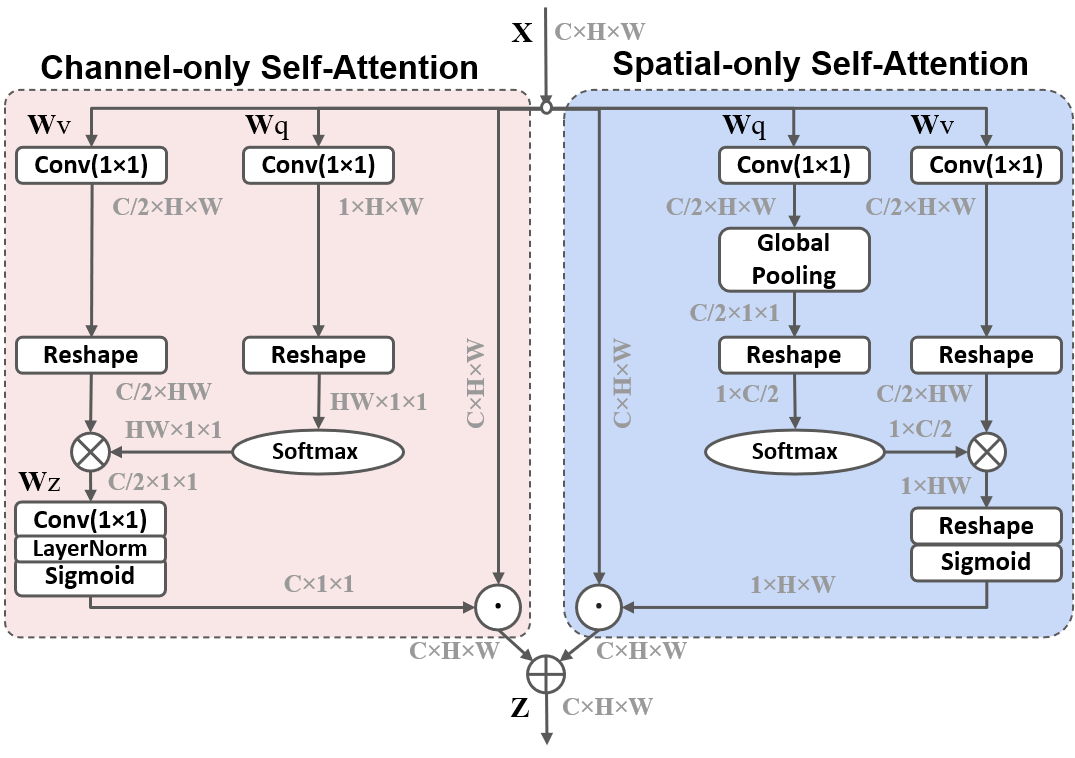}
\includegraphics[width=.99\linewidth]{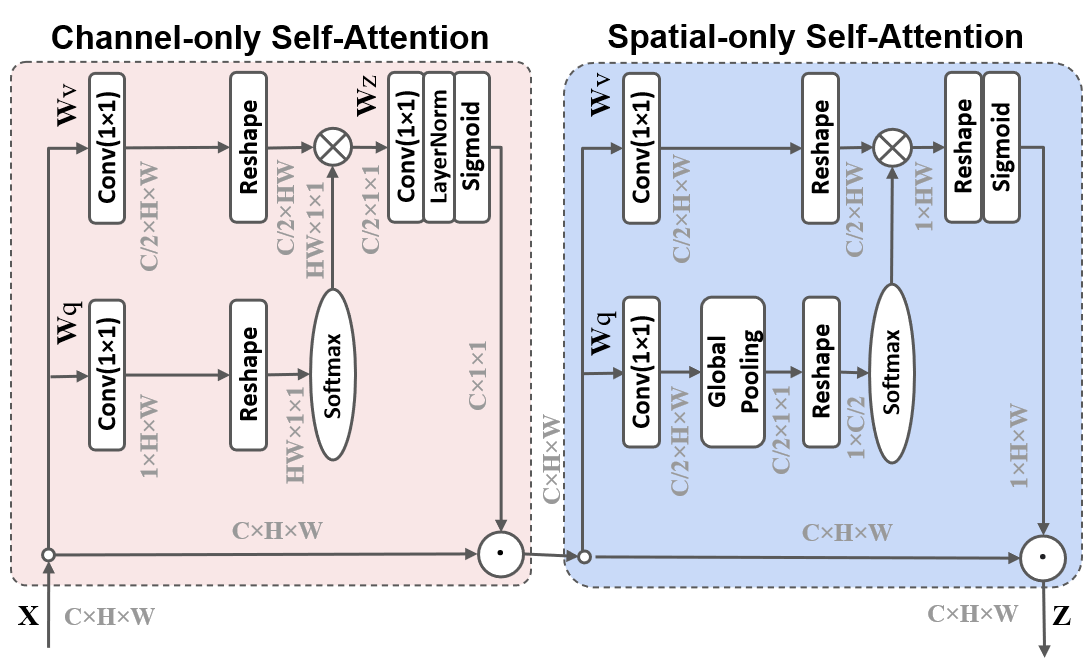}
\caption{The Polarized Self-Attention (PSA) block under \textbf{(upper)} the parallel layout,  and \textbf{(lower)} the sequential layout. }
  \label{fig:PSAlayer}
\end{figure}

Our solution to the above challenges is to conduct ``polarized filtering" in attention computation. A self-attention block operates on an input tensor $\X$ to highlight or suppress features, which is very much like optical lenses filtering the light. In photography, there are always random lights in transverse directions that produce glares/reflections. Polarized filtering, by only allowing the light pass orthogonal to the transverse direction, can potentially improve the contrast of the photo. Due to the loss of total intensity, the light after filtering usually has a small dynamic range, therefore needs a additional boost, e.g. by High Dynamic Range (HDR), to recover the details of the original scene. 

We borrow the key factors of photography, and propose the Polarized Self-Attention (PSA) mechanism: (1) Filtering: completely collapse features in one direction while preserving high-resolution in its orthogonal direction;  (2) HDR: increase the dynamic range of attention by Softmax normalization at the bottleneck tensor (smallest feature tensor in attention block), followed by tone-mapping with the Sigmoid function. Formally, we instantiate the PSA mechanism as a PSA block below (also see diagram in Figure~\ref{fig:PSAlayer}):

\textbf{Channel-only branch $A^{ch}(\X)\in\Re^{C\times 1 \times 1}$:}  
\begin{equation}
    \A^{ch}(\X)= F_{SG}\Big[\W_{z|\theta_1}\Big((\sigma_1(\W_{v}(\X)) \times F_{SM}(\sigma_2(\W_{q}(\X)))\Big)\Big],
    \label{eqn:CHA}
\end{equation}
where $\W_{q}, \W_{v}$ and $\W_{z}$ are $1\times 1$ convolution layers respectively, $\sigma_1$ and $\sigma_2$ are two tensor reshape operators, and $F_{SM}(\cdot)$ is a SoftMax operator and "$\times$" is the matrix dot-product operation $F_{SM}(\X)= \sum_{j=1}^{N_p}{\frac{e^{x_j}}{\sum_{m=1}^{N_p}{e^{x_m}}}{x_j}}$. The internal number of channels, between $\W_{v}|\W_{q}$ and $\W_{z}$, is $C/2$. The output of channel-only branch is $\Z^{ch}=\A^{ch}(\X) \odot^{ch} \X \in \Re^{C\times H \times W}$, where $\odot^{ch}$ is a channel-wise multiplication operator. 

 \textbf{Spatial-only branch $A^{sp}(\X)\in\Re^{1\times H \times W}$:}  
\begin{equation}
    \A^{sp}(\X)= F_{SG}\Big[\sigma_3\Big(
    F_{SM}(\sigma_1(F_{GP}(\W_{q}(\X)))) 
    \times 
    \sigma_2(\W_{v}(\X))
    \Big)\Big],
    \label{eqn:SPA}
\end{equation}
where $\W_{q}$ and $\W_{v}$ are standard $1\times 1$ convolution layers respectively, $\theta_2$ is an intermediate parameter for these channel convolutions, and $\sigma_1$, $\sigma_2$ and $\sigma_3$ are three tensor reshape operators, and $F_{SM}(\cdot)$ is the SoftMax operator. $F_{GP}(\cdot)$ is a global pooling operator $ F_{GP}(\X)= \frac{1}{H\times W} \sum_{i=1}^{H} \sum_{j=1}^{W} X(:,i,j)$, and $\times$ is the matrix dot-product operation. The output of spatial-only branch is $\Z^{sp}=\A^{sp}(\X) \odot^{sp} \X \in \Re^{C\times H \times W}$, where $\odot^{sp}$ is a spatial-wise multiplication operator. 

\textbf{Composition:} The outputs of above two branches are composed either under the parallel layout
\begin{eqnarray}
  PSA_p(\X)&=&  \Z^{ch}+\Z^{sp}\\
\nonumber &=&\A^{ch}(\X) \odot^{ch} \X + \A^{sp}(\X) \odot^{sp} \X,
\label{eqn:PSA_p}
\end{eqnarray}
or under the sequential layout
\begin{eqnarray}
 PSA_s(\X)&=& \Z^{sp}(\Z^{ch})\\
\nonumber &=& \A^{sp}(\A^{ch}(\X) \odot^{ch} \X) \odot^{sp} \A^{ch}(\X) \odot^{ch} \X.
\label{eqn:PSA_s}
\end{eqnarray}
where "+" is the element-wise addition operator.

\textbf{Relation of PSA to other Self-Attentions:} We add PSA to Table~\ref{table:Analysis} and make the following observations: 
\begin{itemize}
    \item \textit{Internal Resolution vs Complexity:} Comparing to existing attention blocks under their top configuration, PSA preserves the highest attention resolution for both the channel ($C/2$)\footnote{$C/2$ is the smallest channel number when PSA produces the best metrics, and is used throughout our experiments.} and spatial ($[W, H]$) dimension. 
    
    Moreover, in our channel-only attention, the Softmax re-weighting is fused with squeeze-excitation leveraging Softmax as the nonlinear activation at the bottleneck tensor of size $C/2 \times W \times H$. The channel numbers $C$-$C/2$-$C$ follow a squeeze-excitation pattern that benefited both GC and SE blocks. Our design conducts \textbf{higher-resolution squeeze-and-excitation} while at comparable computation complexity of the GC block. 
    
    Our spatial-only attention not only keeps the full $[W,H]$ spatial resolution, but also internally keeps $2\times C \times C/2$ learnable parameters in $W_q$ and $W_v$ for the nonlinear Softmax re-weighting, which is more powerful structure than existing blocks. For instance, the spatial-only attention in CBAM is parameterized by a $7\times 7 \times 2$ convolution (a linear operator), and EA learns $ C \times d_{k}+ C \times d_{v}$ parameters for linear re-weighting ($d_{k}, d_{v} \ll C$ ). 

    \item \textit{Output Distribution/Non-linearity.} Both the PSA channel-only and spatial-only branches use a Softmax-Sigmoid composition. Considering the Softmax-Sigmoid composition as a probability distribution function, both the multi-mode Gaussian maps (keypoint heatmaps) and the piece-wise Binomial maps (segmentation masks) can be approximated upon linear transformations, i.e. $1\times 1$ convolutions in PSA. We therefore expect the non-linearity could fully leverage the high resolution information preserved within in PSA attention branches.
    % \begin{eqnarray}
    %  softmax(x_{i})&=& \frac{exp(x_{i})}{\Sigma_{j=1}^{C}exp(x_j)},\\
    %  sigmoid(x_{i})&=& \frac{1}{1+exp(-x_{i})}, \\
    % && for i=1,\cdots, C. 
    %  \end{eqnarray}
  \end{itemize}

% (Distribution-Aware Coordinate Representation for Human Pose Estimation)
% (Coordinate Attention for Efficient Mobile Network Design)

\section{Experiments}
\textbf{Implementation details.} For any baseline networks with the bottleneck or basic residual blocks, such as ResNet and HRnet, we add PSAs after the first $3\times3$ convolution in every residual blocks, respectively. For 2D pose estimation, we kept the same training strategy and hyper-parameters as the baseline networks. For semantic segmentation, we added a warming-up training phase of 5000 iterations, stretched the total training iteration by $30\%$, and kept all the rest training strategy and hyper-parameters of the baseline networks. Empirically, these changes allow PSA to train smoothly on semantic segmentation. 
%   (see Fig.~\ref{fig:PSAinBlock} ). 
% \begin{figure}[!htb]
% \centering
% \includegraphics[width=.90\linewidth]{Images/FigPSAinBlock.png}
% \caption{Adding PSA blocks into the bottleneck and basic blocks of ResNet varients. }
%   \label{fig:PSAinBlock}
% \end{figure}
%\textcolor{red}{Note that, with the same changes of training tricks, all the vanilla networks did not show noticeable performance gain.} 

\subsection{PSA vs. Baselines}
We first add PSA blocks to standard baseline networks of the following tasks.

\begin{table*}[!htb]
\centering
\fontsize{7}{8}\selectfont
\setlength{\tabcolsep}{4.6pt}
\begin{tabular}{l|c|c|cccccc|c|c}
\hline
Method       & Backbone  & ImageNet Pretrain    & $\mathrm{AP}$ $\uparrow$  & $\mathrm{AP_{50}}$$\uparrow$ & $\mathrm{AP_{75}}$$\uparrow$ & $\mathrm{AP_{M}}$ $\uparrow$ & $\mathrm{AP_{L}}$$\uparrow$  & $\mathrm{AR}$ $\uparrow$ & Flops  & mPara  \\ \hline \hline
%\textbf{\textit{State-of-the-Arts}} &  & & & & & &&\\
Simple-Baseline~\cite{Xiao18} & Res50 & Y &  72.2 &89.3 &78.9 & 68.1 &79.7 &77.6 &20.0G&34.0M\\
\textbf{+PSA}    & Res50   &N   &  \textbf{76.5(+4.3)} & \textbf{93.6} & \textbf{83.6} & \textbf{73.2} & \textbf{81.0} &  \textbf{79.0} & 20.9G & 36.1M\\
\hline
Simple-Baseline~\cite{Xiao18} & Res152 & Y &  74.3 &89.6 &81.1 &70.5 &81.6 &79.7 &35.3G&68.6M\\
\textbf{+PSA}      & Res152 & N    &  \textbf{78.0(+3.7)} & \textbf{93.6} & \textbf{84.8} & \textbf{75.2} & \textbf{82.3} &  \textbf{80.5}  &37.5G & 75.2M  \\
\hline
HRNet~\cite{Sun2019} & HRNet-W32 & Y &   75.8 &90.6  &82.5 & 72.0 & 82.7 &  80.9 &16.0G&28.5M\\
\textbf{+PSA}   & HRNet-W32 & Y& \textbf{78.7(+2.9)} & \textbf{93.6} & \textbf{85.9} & \textbf{75.6} & \textbf{83.5} &  \textbf{81.1}  &17.1G & 31.4M\\ 
\hline
HRNet~\cite{Sun2019} & HRNet-W48 & Y &   76.3 &90.8 &82.9 &72.3 &83.4 &81.2 &32.9G&63.6M\\
\textbf{+PSA}      & HRNet-W48 & Y    &  \textbf{78.9(+2.6)} & \textbf{93.6} & \textbf{85.7} & \textbf{75.8} & \textbf{83.8} &  \textbf{81.4}  &35.2G&70.0M\\ 
%RSN \cite{xxx} & 4$\times$RSN-50 & N &   0.796 &  0.925 & 0.858 & 0.755 & 0.832&  0.842 &65.9G&-\\
%RSN \cite{xxx} & 4$\times$RSN-50$^{+}$ & Y &   0.792 &  0.944 & 0.871 & 0.761 & 0.838 &  0.841 &68.6G&-\\
%\hline
\hline
\end{tabular}
\caption{ PSA vs. Baselines for top-down human pose estimation on the MS-COCO val2017 dataset. All results were computed with an human detector~\cite{Xiao18} of 56.4 AP on COCO val2017 dataset. All detected human image patches were resized to $384\times 288$. }
\label{table:COCO_Pose}
\end{table*}

\begin{table}[!htb]
\centering
\fontsize{7}{8}\selectfont
\setlength{\tabcolsep}{5.6pt}
\begin{tabular}{l|c|c|c|c}
\hline
Method       & Backbone & mIoU $\uparrow$ & Flops  & mPara  \\ \hline \hline
DeepLabV3Plus~\cite{Chen17} & MobileNet  &71.1 &16.9G & 5.22M\\
\textbf{+PSA}      & MobileNet     & \textbf{73.7(+2.6)} & 17.1G & 5.22M  \\
\hline
DeepLabV3Plus~\cite{Chen17} & Res50  &77.2 &62.5G &39.8M\\
\textbf{+PSA}     & Res50   & \textbf{79.0(+1.8)} & 65.2G & 42.3M  \\
\hline
DeepLabV3Plus~\cite{Chen17} & Res101 &78.3 &83.2G &58.8M\\
\textbf{+PSA}      & Res101    & \textbf{80.3(+2.0)} & 87.7G & 63.5M  \\
\hline 
\end{tabular}
\caption{ PSA vs. Baselines for semantic segmentation on the Pascal VOC2012 Aug database. }
\label{table:VOC_Semantic}
\end{table}

\textbf{Top-Down 2D Human Pose Estimation}: Among the DCNN approaches for 2D human pose estimation, the top-down approaches generally dominate the top metrics. This top-down pipeline consists of a person bounding box detector and a keypoint heatmap regressor. Specifically, we use the pipelines in \cite{Xiao18} and \cite{Sun2019} as our baselines. An input image is first processed by a human detector~\cite{Xiao18} of $56.4$AP (Average Precision)  on MS-COCO val2017 dataset~\cite{Lin2014COCO}. Then all the detected human image patches are cropped from the input image and resized to $384\times 288$. Finally, the $384\times 288$ image patches are used for keypoint heatmap regression by a single person pose estimator. The output heatmap size is $96 \times 72$.

We add PSA on Simple-Baseline~\cite{Xiao18} with the Resnet50/152 backbones and HRnet~\cite{Sun2019} with the HRnet-w32/w48 backbones. The results on MS-COCO val2017 are shown in Table~\ref{table:COCO_Pose}. PSA boosts all the baseline networks by $2.6$ to $4.3$ AP with minor overheads of computation (Flops) and the number of parameters(mPara). Even without ImageNet pre-training, PSA with ``Res50'' backbone gets $76.5$ AP, which is not only $4.3$ better than Simple-Baseline with Resnet50 backbone, but also better than Simple-Baseline even with Resnet152 backbone. A similar benefit is also observed on PSA with HRNet-W32 backbone outperforms the baseline with ``HR-w48'' backbone. This giant performance gains of PAS and the small overheads make PSA+HRNet-W32 the most cost-effective model among all models in Table~\ref{table:COCO_Pose}.

\textbf{Semantic Segmentation.}  This task maps an input image to a stack of segmentation masks, one output mask for one semantic class. In Table~\ref{table:VOC_Semantic}, we compare PSA with the DeepLabV3Plus~\cite{Chen17} baseline on the Pascal VOC2012 Aug{~\cite{Everingham15}} (21 classes, input image size $513 \times 513$, output mask size $513 \times 513$). PSA boosts all the baseline networks by $1.8$ to $2.6$mIoU(mean Intersection over Union) with minor overheads of computation (Flops) and the number of parameters (mPara).  PSA with ``Res50'' backbone got $79.0$ mIoU, which is not only $1.8$ better than the DeepLabV3Plus with the Resnet50 backbone, but also better than DeepLabV3Plus even with Resnet101.

\subsection{Comparing with State-of-the Arts}
\begin{figure*}[!htb]
\centering
\includegraphics[width=\textwidth]{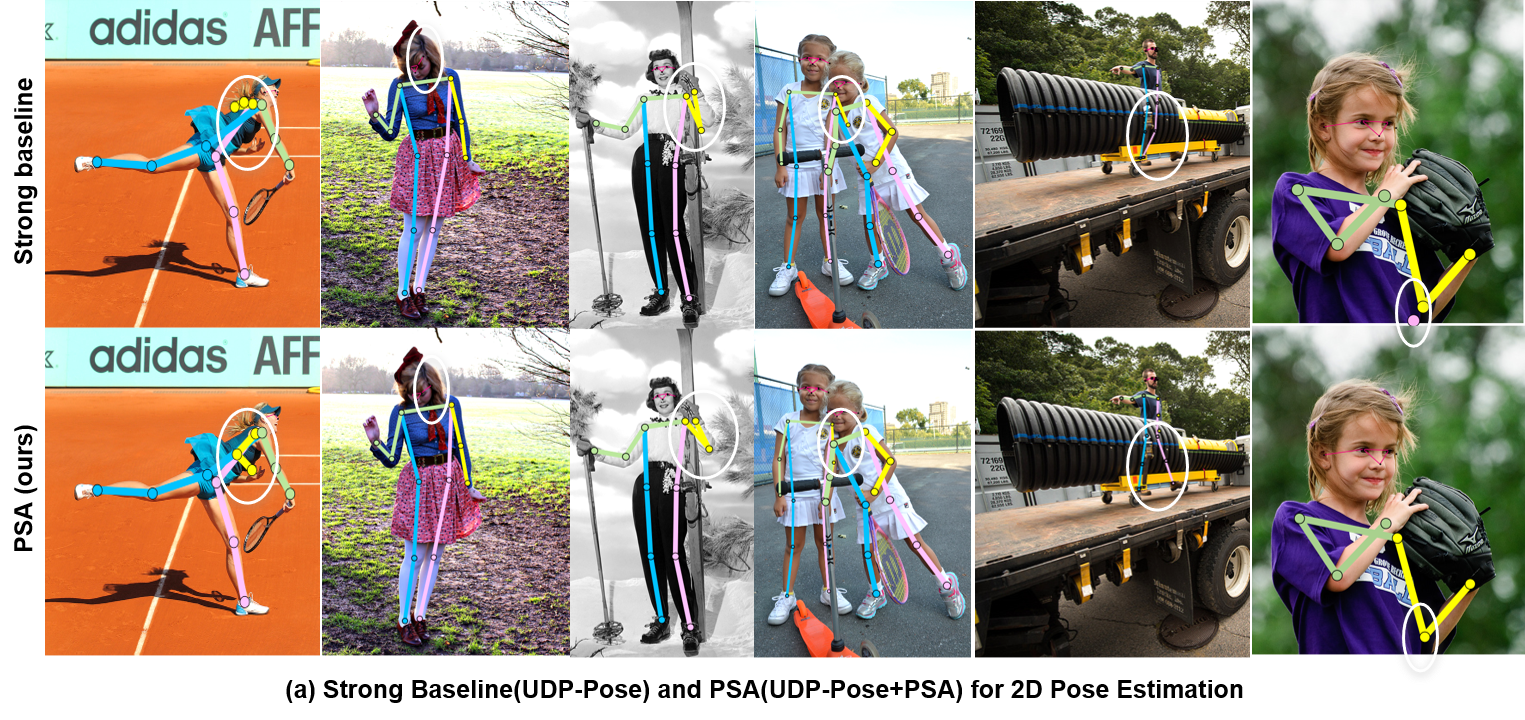}
\includegraphics[width=\textwidth]{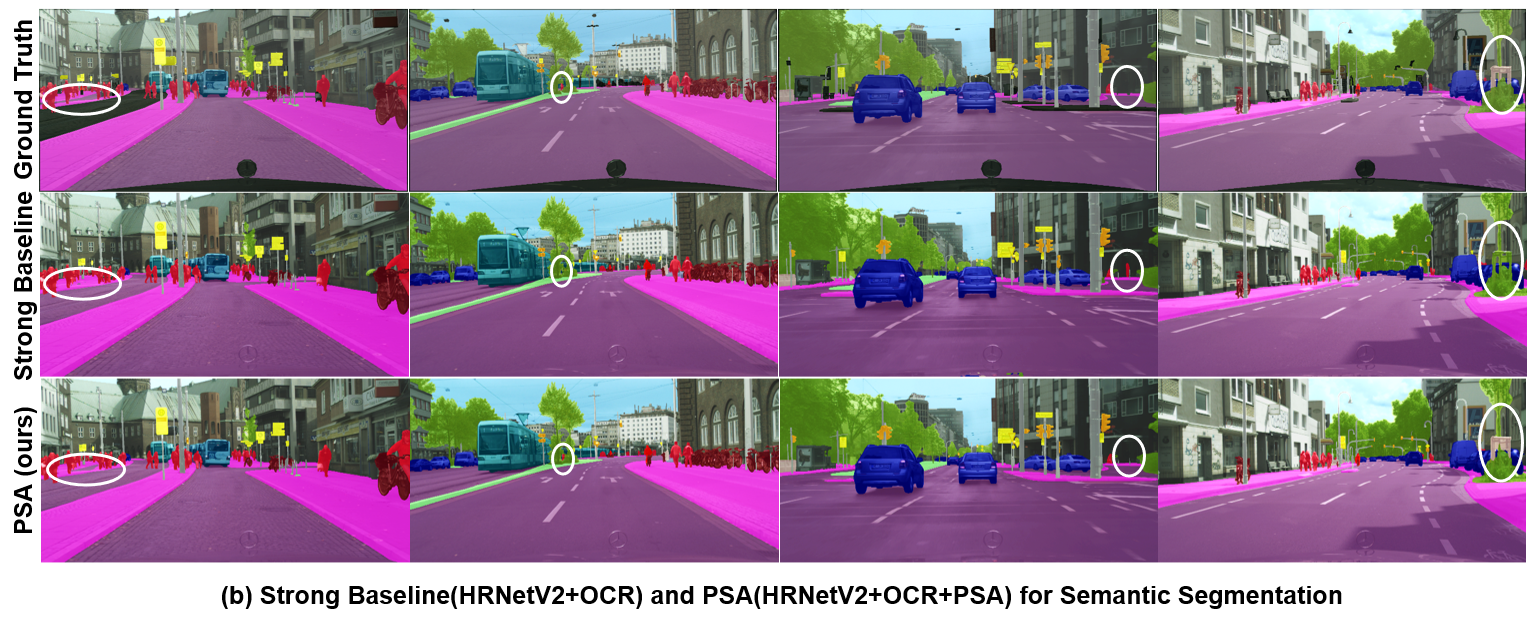}
\caption{Qualitative comparison of PSA(ours) and \textbf{\textsl{Strong Baselines}}: (a) Human Pose Estimation(UDP-Pose, Table~\ref{table:CocoSOTA}) and (b) Semantic segmentation(HRNetV2-OCR, Table~\ref{table:CityscapesSOTA} ). The white eclipses highlight the fine-grained details that PSAs outperform the strong baselines. }
  \label{fig:QualitativeStrong}
\end{figure*}

We then apply PSA to the current state-of-the-arts of above tasks.
\textbf{Top-down 2D Human Pose Estimation.} To our knowledge, the current state-of-the-art results by single models were achieved by UDP-HRnet with 65.1mAP bbox detector on the MS-COCO keypoint testdev set. In Table~\ref{table:CocoSOTA}, we add PSA to the UDP-Pose with HRnet-W48 backbone and achieve a new state-of-the-art AP of $79.5$. PSA boosts UDP-Pose (baseline) by $1.7$ points (see Figure~\ref{fig:QualitativeStrong} (a) for their qualitative comparison). 

Note that there is only a subtle metric difference between the parallel (p) and sequential(s) layout of PSA. We believe this partially validate that our design of the channel-only and spatial-only attention blocks has exhausted the representation power along the channel and spatial dimension. 

\begin{table*}[t]
\centering
\fontsize{7}{8}\selectfont
\begin{tabular}{l|c|c|c|ccccc|c|c}
\hline
Method       & Backbone   & Input Size    & $\mathrm{AP}$   & $\mathrm{AP_{50}}$ & $\mathrm{AP_{75}}$ & $\mathrm{AP_{M}}$  & $\mathrm{AP_{L}}$  & $\mathrm{AR}$ & Flops& mPara \\ \hline \hline
% \textbf{\textit{Top Down}} & & & & & & & &&\\
8-stage Hourglass~\cite{Newell16} & 8-stage Hourglass & $256\times192$ & 66.9 & - &- & - & 
- &  - & 14.3G & 25.1M\\
CPN \cite{Chen2018} & ResNet50 & $256\times192$ & 68.6 & - & - & - & - & - & 6.2G & 27.0M\\
CPN + OHKM  \cite{Chen2018} & ResNet50 & $256\times192$ & 69.4 & - & - & - & - & - & 6.2G & 27.0M\\
SimpleBaseline \cite{Xiao18} & ResNet50 & $256\times192$ & 70.4 & 88.6 &78.3 & 67.1 & 77.2 & 76.3 & 8.90G &34.0M\\
SimpleBaseline \cite{Xiao18} & ResNet101 & $256\times192$ & 71.4 & 89.3 & 79.3 & 68.1 & 78.1 & 77.1 &12.4G &53.0M\\
SimpleBaseline \cite{Xiao18} & ResNet152 & $256\times192$ & 72.0 & 89.3 &79.8 & 68.7 & 78.9 & 77.8 &15.7G &72.0M\\
HRNet-W32 \cite{Sun2019} & HRNet & $256\times192$ & 74.4 & 90.5 &81.9 & 70.8 & 81.0 & 78.9 &7.10G &28.9M\\
HRNet-W48 \cite{Sun2019} & HRNet & $256\times192$ & 75.1 & 90.6 &82.2 & 71.5 & 81.8 &  80.4 &14.6G&63.6M\\
Dark-Pose \cite{Zhang2020} & HRNet-W32 & $256\times192$ & 75.6 & 90.5 &82.1 & 71.8 & 82.8 & 80.8 &7.1G &28.5M\\
UDP-Pose \cite{Huang2020} & HRNet-W48 & $256\times192$ & 77.2 & 91.8 &83.7 & 73.8 & 83.7 & 82.0 &14.7G &63.8M\\
\hline
% \textbf{\textit{One-stage}} & & & & & & & &&\\
SimpleBaseline \cite{Xiao18} & ResNet152 & $384\times288$ & 74.3 & 89.6 & 81.1 & 70.5 & 79.7 & 79.7 &35.6G &68.6M\\
HRNet-W32 \cite{Sun2019} & HRNet & $384\times288$ & 75.8 & 90.6 & 82.7 & 71.9 & 82.8 & 81.0 &16.0G &28.5M\\
HRNet-W48 \cite{Sun2019} & HRNet & $384\times288$ & 76.3 & 90.8 & 82.9 & 72.3 & 83.4 & 81.2 &32.9G &63.6M\\
Dark-Pose \cite{Zhang2020} & HRNet-W48 & $384\times288$ & 76.8 &90.6 & 83.2 & 72.8 & 84.0 & 81.7 &32.9G &63.6M\\
UDP-Pose \cite{Huang2020} & HRNet-W48 & $384\times288$ &76.2& 92.5& 83.6& 72.5& 82.4& 81.1&33.0G &63.8M\\
UDP-Pose \cite{Huang2020} (\textbf{\textsl{Strong Baseline}}) & HRNet-W48 & $384\times288$ & 77.8 & 92.0 &84.3 &74.2  &\textbf{84.5}  & \textbf{82.5} &33.0G &63.8M\\
\hline
\textbf{\textit{Ours}} & & & & & & & & && \\
UDP-Pose-PSA(p)      & HRNet-W48   & $256\times192$  & 78.9 & \textbf{93.6} & \textbf{85.8} & \textbf{76.1} & 83.6 &  81.4 & 15.7G & 70.1M  \\
UDP-Pose-PSA(p)      & HRNet-W48  & $384\times288$ & \textbf{79.5} & \textbf{93.6} & \textbf{85.9} & \textbf{76.3} & \textbf{84.3} & 81.9 &35.4G &70.1M  \\
UDP-Pose-PSA(s)      & HRNet-W48  & $384\times288$ & \textbf{79.4} & \textbf{93.6} & \textbf{85.8} & \textbf{76.1} & 84.1 & 81.7 &35.4G &69.1M  \\
 \hline
\end{tabular}
\caption{ Comparison with State-of-the-Art top-down 2D pose estimation approaches on the MS-COCO keypoint testdev set. Note that only \cite{Huang2020}\textbf{\textsl{Strong Baseline}} used extra training data.}
\label{table:CocoSOTA}
\end{table*}

\textbf{Semantic Segmentation.} To our knowledge, the current state-of-the-art results by single models were produced by HRNet-OCR(MA)~\cite{Tao2020} on the Cityscapes validation set {~\cite{Cordts16}}(19 classes, input image size $1024 \times 2048$, output mask size $1024\times 2048$). In Table~\ref{table:CityscapesSOTA}, we add PSA to the basic configuration of HRNet-OCR and achieve the new state-of-the-arts mIoU of $86.95$. PSA boosts HRNet-OCR (strong baseline) by $2$ points(see Figure~\ref{fig:QualitativeStrong} (b) for their qualitative comparison). Again that there is only a subtle metric difference between the PSA results under the parallel(p) layout and the sequential(s) layout.

\begin{table*}[!htb]
\centering
\fontsize{7}{8}\selectfont
\setlength{\tabcolsep}{4.6pt}
\begin{tabular}{l|c|cccc}
\hline
Method       & Backbone   & mIoU  & iIoU cla. & IoU cat. & iIoU cat.  \\ \hline \hline
GridNet~\cite{Fourure2017} & -  &  69.5 &44.1 &87.9 &71.1 \\
LRR-4x & -    & 69.7 & 48.0 & 88.2 & 74.7   \\
DeepLab~\cite{Chen17} & D-ResNet-101 & 70.4 &42.6 &86.4 &67.7\\
LC    & - & 71.1 & - & - & -  \\
Piecewise~\cite{Lin2016}  & VGG-16 & 71.6 & 51.7 & 87.3 & 74.1 \\
FRRN~\cite{Pohlen2017}     & - & 71.8 & 45.5 & 88.9 & 75.1   \\
RefineNet~\cite{Lin2017} & ResNet-101 & 73.6 & 47.2 & 87.9 & 70.6   \\
PEARL~\cite{Jin2017} & D-ResNet-101 & 75.4 & 51.6 & 89.2 & 75.1   \\
DSSPN~\cite{Liang2018} & D-ResNet-101 & 76.6 & 56.2 & 89.6 & 77.8   \\
LKM~\cite{Peng2017}  & ResNet-152 & 76.9 & - & - & -   \\
DUC-HDC~\cite{Wang2018} & - & 77.6 & 53.6 & 90.1 & 75.2   \\
SAC~\cite{Zhang2017_sac}  & D-ResNet-101 & 78.1 & - & - & -  \\
DepthSeg~\cite{Kong2018}  & D-ResNet-101 & 78.2 & - & - & - \\
ResNet38~\cite{Wu2019} & WResNet-38 & 78.4 & 59.1 & 90.9 & 78.1 \\
BiSeNet~\cite{Yu2018_bisenet} & ResNet-101 & 78.9 & - & - & -   \\
DFN~\cite{Yu2018} & ResNet-101 & 79.3 & - & - & -   \\
PSANet~\cite{Zhao2018} & D-ResNet-101 & 80.1 & - & - & -   \\
PADNet~\cite{Xu2018}  & D-ResNet-101 & 80.3 & 58.8 & 90.8 & 78.5   \\
CFNet~\cite{Zhang2019}  & D-ResNet-101 & 79.6 & - & - & -   \\
Auto-DeepLab~\cite{Liu2019} & - & 80.4 & - & - & -   \\
DenseASPP~\cite{Zhao2017}  & WDenseNet-161 & 80.6 & 59.1 & 90.9 & 78.1   \\
SVCNet~\cite{Ding2019}  & ResNet-101 & 81.0 & - & - & -   \\
ANN~\cite{Zhu19_ann}  & D-ResNet-101 & 81.3 & - & - & -   \\
CCNet~\cite{Huang2019} & D-ResNet-101 & 81.4 & - & - & -  \\
DANet~\cite{Fu2019} & D-ResNet-101 & 81.5 & - & - & -  \\
HRNetV2~\cite{Wang2020} & HRNetV2-W48 & 81.6 & 61.8 & 92.1 & 82.2 \\
% HRNetV2+OCR~\cite{} & HRNetV2-W48 & 82.5 & 61.7 & 92.1 & 81.6   \\
HRNetV2+OCR~\cite{Yuan2020} & HRNetV2-W48 & 84.9 & - & - & -   \\
HRNetV2+OCR(MA)~\cite{Tao2020}  (\textbf{\textsl{Strong Baseline}}) & HRNetV2-W48 & 85.4 & - & - & -   \\
% HRNetV2+OCR(AL)~\cite{} & HRNetV2-W48 & 86.0 & - & - & -   \\
% HRNetV2+OCR(MA+AL)~\cite{} & HRNetV2-W48 & 86.3 & - & - & -   \\
\hline
\textbf{\textit{Ours}}  &  & & & & \\
HRNetV2-OCR+PSA(p) & HRNetV2-W48 & \textbf{86.95}  &\textbf{71.6}  &\textbf{92.8}   &\textbf{85.0} \\
HRNetV2-OCR+PSA(s) & HRNetV2-W48 & \textbf{86.72}  &\textbf{71.3}  &92.3   &82.8 \\
\hline
\end{tabular}
\caption{Comparison with State-of-the-Art semantic segmentation approaches on the Cityscapes validation set. }
\label{table:CityscapesSOTA}
\end{table*}

\subsection{Ablation Study}

In Table~\ref{table:Ablation}, we conduct an ablation study of PSA configurations on Simple-Baseline(Resnet50)~\cite{Xiao18} and compare PSAs with other related self-attention methods. All the overheads, such as Flops, mPara, inference GPU memory("Mem."), and inference time ("Time")) are inference costs of \textbf{one sample}. To reduce the randomness in CUDA and Pytorch scheduling, we ran inference on MS-COCO val2017 using 4 TITAN RTX GPUs, batchsize 128 (batchsize 32/GPU), and averaged over the number of samples.

From the results of "PSA ablation" in Table~\ref{table:Ablation}, we observe that \textbf{(1)} the channel-only block ($\A^{ch}$) outperform spacial-only attention ($\A^{sp}$), but can be further boosted by their parallel ([$\A^{ch}|\A^{sp}$]) or sequential ($\A^{sp}(\A^{ch})$) compositions; \textbf{(2)} The parallel ([$\A^{ch}|\A^{sp}$]) or sequential ($\A^{sp}(\A^{ch})$) compositions has similar AP, Flops, mPara, inference memory(Mem.), and inference (Time.).

From the results of "related self-attention methods", we observe that \textbf{(1)} the NL block costs the most memory while produces the least boost ($2.3$AP) over the baseline, indicating that NL is highly redundant. \textbf{(2)} The channel-only attention GC is better than SE since it includes SE. GC is even better than channel+spatial attention CBAM because the inner-product-based attention mechanism in GC is more powerful than the convolution/MLP-based CBAM. \textbf{(3)} PSA $\A^{ch}$ is the best channel-only attention block over GC and SE. We believe PSA benefits from its highest channel resolution ($C/2$) and its output design. \textbf{(4)} The channel+spatial attention CBAM with a relatively early design is still better than the channel-only attention SE. \textbf{(5)} Under the same sequential layout of spatial and channel attention, PSA is significantly better than CBAM. Finally, \textbf{(6)} At similar overheads, both the parallel and sequential PSAs are better than the compared blocks.

\begin{table*}[!htb]
\centering
\fontsize{7}{8}\selectfont
\setlength{\tabcolsep}{1.6pt}
\begin{tabular}{l||cccccc|c|c|c|c|}
\hline
Method       &   $\mathrm{AP}$ $\uparrow$  & $\mathrm{AP_{50}}$$\uparrow$ & $\mathrm{AP_{75}}$$\uparrow$ & $\mathrm{AP_{M}}$ $\uparrow$ & $\mathrm{AP_{L}}$$\uparrow$  & $\mathrm{AR}$ $\uparrow$ & Flops $\downarrow$ & mPara $\downarrow$ & Mem.(MiB) $\downarrow$& Time(ms)$\downarrow$\\ \hline \hline
%\textbf{\textit{State-of-the-Arts}} &  & & & & & &&\\
Simple-Baseline(ResNet50)~\cite{Xiao18} &72.2 &89.3 &78.9 &68.1 &79.7 &77.6 &\textbf{20.0}G&\textbf{34.0}M& 1.43 & \textbf{2.56}\\
\hline
\textbf{\textit{PSA ablation}}    &     &   &   &   &   &    & & & &\\
+$\A^{ch}$    &  76.3(+4.1) & 92.6 & \textbf{83.6} & 73.0 & 80.8 & 78.9   &20.4G &35.3M &  1.49  & 2.58 \\
+$\A^{sp}$    &  75.0(+2.8) & 92.6 & 81.6 & 71.5 & 80.2 & 77.7   &20.7G &35.3M& 1.45 & 2.63  \\
+[$\A^{ch}|\A^{sp}$] (\textbf{PSA(p)})  & \textbf{76.5(+4.3)} & \textbf{93.6} & \textbf{83.6} & \textbf{73.2} & 81.0 &  \textbf{79.0}  &20.9G &36.5M & 1.54 & 2.70 \\
+$\A^{sp}(\A^{ch})$  (\textbf{PSA(s)})  & \textbf{76.6}\textbf{(+4.4)} & \textbf{93.6} & \textbf{83.6} & \textbf{73.2} & \textbf{81.2} & \textbf{79.1}   &20.9G &36.5M & 1.52 & 2.71 \\
\hline
\textbf{\textit{Related self-attention methods}}    &     &   &   &   &   &    & & && \\
 +$\A$ (NL~\cite{Wang18nonlocal}) &  74.5(+2.3) & 92.6 & 81.5 & 70.9 & 79.9 & 77.3   & 21.1G & 36.5M& 10.97 & 2.76 \\
%  +$LowRank(\A)$ (EA~\cite{})   &  0.\textbf{(+0.)} &0  & 0 & 0 & 0 &  0  & 0G & 0.0M&&\\ 
 +$\A^{ch}$ (GC~\cite{Cao19})    &  76.1(+3.9) & 92.6 & 82.7 & 72.9 & 80.9 &  78.7  &20.2G &34.3M & 1.47 & 2.69\\
 +$\A^{ch}$ (SE~\cite{Hu18})    &  75.7(+3.5) & \textbf{93.6} & 82.6 & 72.4 & 80.8 & 78.3   &20.2G &34.2M & \textbf{1.29} & 2.94 \\
 +$\A^{sp}(\A^{ch})$ (CBAM~\cite{Sanghyun2018cbam})   & 75.9(+3.7) & 92.6 & 82.7 & 72.9 & 80.7 &  78.7  & 20.2G & 34.3M& 1.49 & 2.96\\ 
  
\hline
\end{tabular}
\caption{Ablation study of PSA and comparison with related attention blocks(human pose estimation on the MS-COCO val2017 dataset with human detector~\cite{Xiao18} of $56.4$AP, input size $384\times 288$.) $\A^{ch}$ denotes channel-only self-attention. $\A^{ch}$ denotes spatial-only self-attention. [$\A^{ch}|\A^{sp}$] denotes the parallel layout of the channel-only and spatial-only self-attention. $\A^{ch}(\A^{sp})$ denotes the sequentially layout. ``Mem" and ``Time" are inference costs of \textbf{one sample}, which are averaged over the val2017 set.}
\label{table:Ablation}
\end{table*}

\section{Conclusion and Future Work}
We presented the Polarized Self-Attention(PSA) block towards high-quality pixel-wise regression. PSA significantly boosts all compared DCNNs for two critical designs (1) keeping high internal resolution in both polarized channel-only and spatial-only attention branches, and (2) incorporating a nonlinear composition that fully leverages the high-resolution information preserved in the PSA branches. PSA can potentially benefit any computer vision tasks with pixel-wise regression. 

It is still not clear how PSA would best benefit pixel-wise regression embedded with the classification and displacement regression in complex DCNN heads, such as those in the instance segmentation, anchor-free object detection and panoptic segmentation tasks. To our knowledge, most existing work with self-attention blocks only inserted blocks in the backbone networks. Our future work is to explore the use of PSAs in DCNN heads.

% It is not clear how PSA would best benefit pixel-wise regression incorporated in complex head DCNN structures with classification and displacement regression, such as in the instance segmentation and panoptic segmentation tasks. The most closely related example we found are the MaskRCNN with EA~\cite{Shen20} and GCnet\footnote{https://github.com/xvjiarui/GCNet} for instance Segmentation. Even with complex head DCNN structures, existing work only inserted self-attention blocks in the backbone. In Table.~\ref{table:COCO_MaskRCNN}, we conduct preliminary comparison by directly replacing GC blocks with PSA.  \textcolor{red}{We found that xxx}. Our future work is to investigate the use of self-attention blocks in different DCNN heads.

% \begin{table*}[!htb]
% \centering
% \fontsize{7}{8}\selectfont
% \setlength{\tabcolsep}{1.6pt}
% \begin{tabular}{l|c|c||c|c||c|c}
% \hline
% Method & Backbone  &   Lr schedule    & Box AP $\uparrow$ & Mask AP $\uparrow$  & Flops $\downarrow$ & mPara $\downarrow$\\ \hline \hline
% \textbf{\textit{Detectron2}} & & &  & & &\\
% Mask R-CNN &R50-FPN       & $3\times$ & 41.0& 37.2 & 0 G& 0 M\\
% Mask R-CNN &X101-32$\times$8d-FPN  & $3\times$ & 44.3& 39.5 & 0 G& 0 M\\
% \hline
% \textbf{\textit{Ours}} & & & &  & & \\
% PSA& PSA-R50-FPN      & $3\times$     & &  & 0 G& 0 M  \\
% PSA& PSA-X101-32$\times$8d-FPN & $3\times$    & &  & 0 G& 0 M  \\
%  \hline
% \end{tabular}
% \caption{Adding PSA to the instance segmentation head of MaskRCNN baseline vs. for instance segmentation on the MS-COCO val2017 dataset.}
% \label{table:COCO_MaskRCNN}
% \end{table*}

{\small
\bibliographystyle{ieee_fullname}
\bibliography{mainbib}
}

\end{document}